\documentclass[sigconf]{acmart}

\setcopyright{none}
\renewcommand\footnotetextcopyrightpermission[1]{}
\usepackage{booktabs}
\usepackage{algorithm}
\usepackage{algpseudocode}
\usepackage{balance}
\usepackage{tikz}
\usetikzlibrary{arrows.meta}
\definecolor{slate}{RGB}{72,94,130}

\begin{document}

\title[The Irreversibility Budget]{The Irreversibility Budget: Fleet-Level Risk Accounting and Admission Control for Agent Operating Systems}

\acmConference[Agentic OS Workshop]{Agentic OS Workshop, SOSP 2026}{September 29, 2026}{Prague, Czechia}

\author{Bardia Mohammadi}
\email{bmohammadi@mpi-sws.org}
\affiliation{%
  \institution{Max Planck Institute for Software Systems}
  \city{Saarbr\"ucken}
  \country{Germany}}

\author{Laurent Bindschaedler}
\email{bindsch@mpi-sws.org}
\affiliation{%
  \institution{Max Planck Institute for Software Systems}
  \city{Saarbr\"ucken}
  \country{Germany}}

\begin{abstract}
Fleets of LLM agents now externalize effects that cannot be fully undone: they move money, deploy code, delete data, and disclose information. Current controls check one effect at a time, so a fleet of individually authorized agents can overdraw its principal's risk under a shared trigger while every local gate stays correct. We propose the \emph{irreversibility budget}, a cumulative account of residual value-at-risk that a trusted runtime maintains for each principal across agents, workflows, and tenants. Treating irreversibility as a first-class resource, the runtime charges each effect its residual loss below the agent and denies the marginal effect once the aggregate would overdraw the budget. Getting the price right is hard, because effects are heterogeneous, adversarially declared, and correlated. We perform a controlled study in which per-effect gates admit fleet-level overdraws of up to $48\times$ the tenant's risk limit while the budget holds every correctly charged run within that limit. Conservative, dependency-aware pricing remains the central open problem for a deployable design.
\end{abstract}

\maketitle
\raggedbottom

\section{Introduction}
Fleets of LLM agents now call tools, update durable state, and externalize effects whose consequences may not be fully reversible. They move money, deploy code, delete data, and disclose information. Managing scarce or dangerous shared quantities is the classic job of an operating system, which measures memory, CPU time, and I/O bandwidth, allocates them to principals, charges them at use, and defends them under contention. Agent operating systems are only beginning to emerge~\cite{Packer2023MemGPT,Mei2025AIOS}, but any of them will inherit this job for a quantity today's runtimes leave unmanaged: the irreversible exposure a fleet creates in the outside world. Unlike memory, it is a residual loss that depends on the effect, its compensation, and how other agents are spending at the same time.

Existing controls do not manage this quantity. They answer local questions: one defers settlement, another validates a task's authority, a third prices a review, and a fourth insures an individual action~\cite{Mohammadi2026Atomix,Chen2026Cordon,Madras2018LearningToDefer,Mozannar2020LearningToDefer,Chen2026AAI,Xu2026TraceEconomic,Hua2026ARS}. None keeps a running balance across agents. Consider, for example, fifty individually capped procurement agents that all watch the same supplier price spike and, each acting rationally, decide to buy at once. Every purchase clears its local cap, yet together they commit a seven-figure position no reviewer approved. Each action is allowed, but the fleet overdraws. Accounting for this before commit is hard: effects differ in how much loss survives them, an agent can misdeclare or split an effect to slip under a cap, and a shared trigger correlates losses that a naive account would simply add up.

\begin{figure*}[t]
\centering
\IfFileExists{figures/IB_Figure.png}{\includegraphics[width=0.68\textwidth]{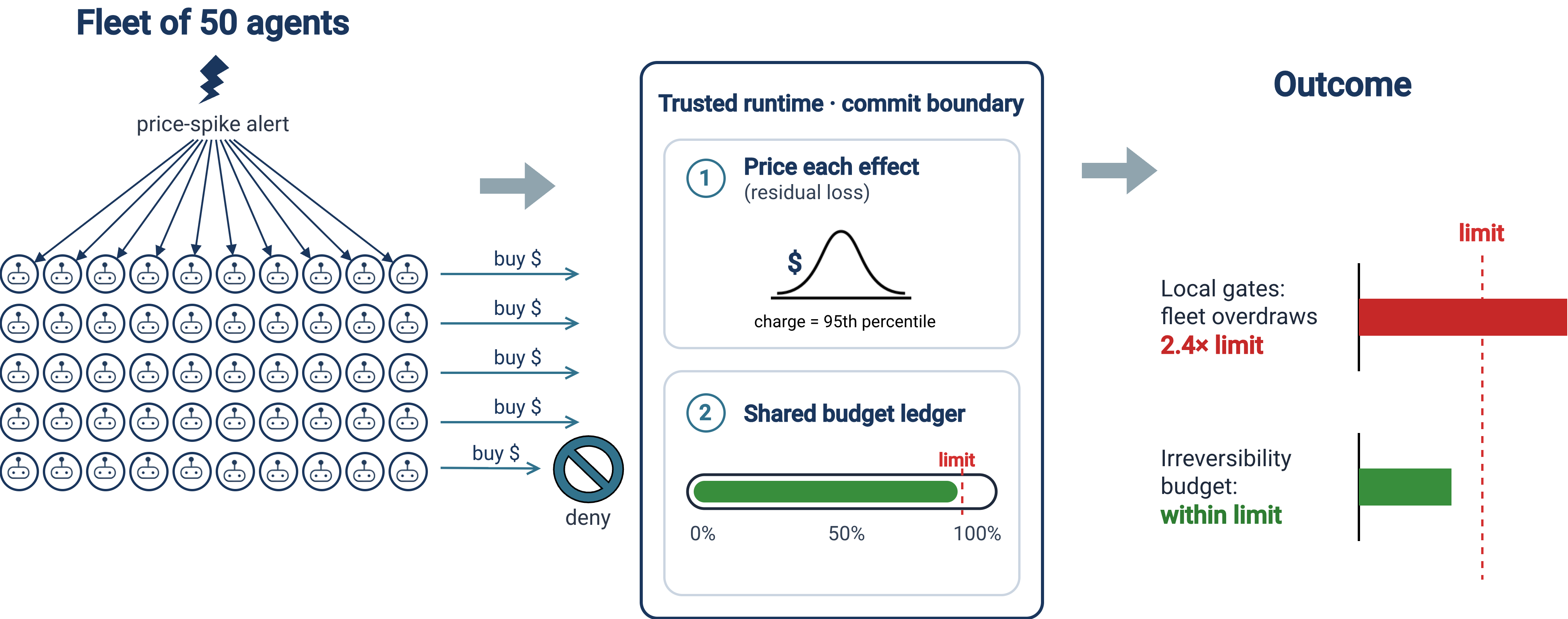}}{%
\resizebox{\textwidth}{!}{%
\begin{tikzpicture}[font=\sffamily\small, >=Latex,
  agent/.style={circle, draw=slate, fill=slate!15, minimum size=2.6mm, inner sep=0pt},
  comp/.style={rounded corners=1.5pt, draw=slate!80, fill=slate!7, align=center, inner sep=4pt, font=\sffamily\footnotesize}]
\node[font=\sffamily\footnotesize\bfseries, text=slate] at (1.35,2.75) {Fleet of 50 agents};
\node[draw=orange!80!black, fill=orange!15, rounded corners=1pt, font=\sffamily\scriptsize, inner sep=2pt] (trig) at (1.35,2.25) {shared trigger: price spike};
\foreach \x in {0,...,9}{ \foreach \y in {0,...,4}{ \node[agent] (a\x\y) at (\x*0.3,\y*0.3) {}; }}
\foreach \x in {0,3,6,9}{ \draw[orange!70!black, opacity=0.45] (trig.south) -- (a\x4); }
\foreach \y in {0,2,4}{ \draw[->, slate!70, thick] (2.85,\y*0.3) -- (4.45,0.6); }
\node[font=\sffamily\scriptsize, text=slate!80] at (3.6,1.15) {buy \$};
\node[draw=slate, thick, rounded corners=3pt, fill=slate!3, minimum width=5.2cm, minimum height=2.95cm] (rt) at (7.1,0.85) {};
\node[font=\sffamily\footnotesize\bfseries, text=slate, anchor=north] at (7.1,2.2) {Trusted runtime $\cdot$ commit boundary};
\node[comp, minimum width=4.4cm] at (7.1,1.5) {\textbf{1}\; price each effect (residual loss)};
\node[comp, minimum width=4.4cm] at (7.1,0.85) {\textbf{2}\; shared budget ledger};
\draw[slate!70] (5.1,0.12) rectangle (9.1,0.42);
\fill[green!60!black] (5.1,0.12) rectangle (8.45,0.42);
\draw[red!75!black, very thick] (9.1,0.02) -- (9.1,0.52);
\node[font=\sffamily\scriptsize, text=red!75!black, anchor=north] at (9.1,0.02) {limit};
\node[circle, draw=red!75!black, fill=red!12, inner sep=1.2pt, font=\sffamily\scriptsize\bfseries, text=red!75!black] at (4.5,0.6) {$\times$};
\node[font=\sffamily\scriptsize, text=red!75!black, anchor=south] at (4.5,0.78) {deny};
\draw[->, slate!70, thick] (9.75,0.85) -- (10.5,0.85);
\node[font=\sffamily\footnotesize\bfseries, text=slate, anchor=south] at (13.0,2.15) {Outcome};
\draw[dashed, gray] (13.0,-0.05) -- (13.0,1.95);
\node[font=\sffamily\scriptsize, text=gray, anchor=south] at (13.0,1.95) {limit};
\fill[red!70!black] (10.65,1.15) rectangle (14.25,1.65);
\node[font=\sffamily\scriptsize, text=white, anchor=west] at (10.8,1.4) {local gates};
\node[font=\sffamily\scriptsize, text=red!70!black, anchor=west] at (14.35,1.4) {$2.4\times$ over};
\fill[green!60!black] (10.65,0.35) rectangle (12.55,0.85);
\node[font=\sffamily\scriptsize, text=white, anchor=west] at (10.8,0.6) {budget};
\node[font=\sffamily\scriptsize, text=green!50!black, anchor=west] at (13.1,0.6) {within limit};
\end{tikzpicture}%
}}
\caption{The irreversibility budget on the running example. Fifty procurement agents share one trigger and each proposes a locally authorized purchase (left). A trusted runtime prices each effect by its residual loss and charges it against one shared ledger, denying the marginal effect once the aggregate would exceed the tenant's limit (center). Per-effect gates overdraw by $2.4\times$, the budget stays within the limit (right).}
\Description{Three-zone diagram. Left: fifty agent icons connected to one shared trigger labeled price spike, each proposing a purchase. Center: a trusted runtime at the commit boundary prices each effect by residual loss and charges it to a shared budget ledger, drawn as a bar filling toward a red limit line, denying the effect that would cross it. Right: outcome bars showing local gates exceeding the limit by 2.4 times while the budget stays within it.}
\label{fig:hero}
\end{figure*}
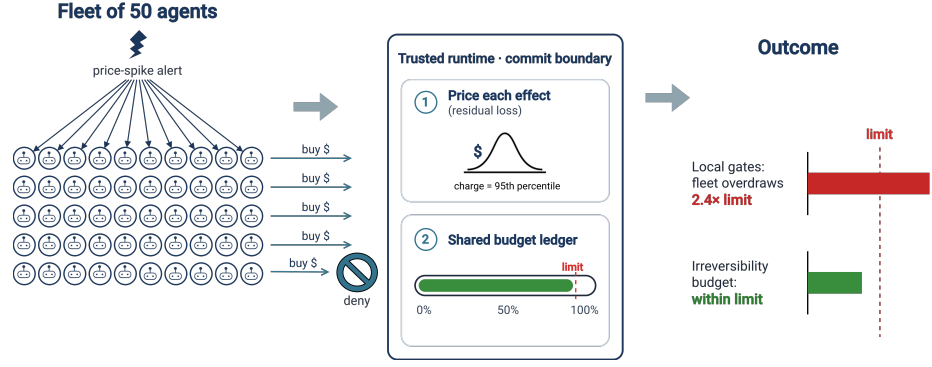

We propose the \emph{irreversibility budget}, a cumulative account of residual value-at-risk that a trusted runtime maintains for every \emph{principal}: an agent, the workflow it runs in, or the tenant that owns them. The key idea is to treat irreversibility as a first-class resource, the way an operating system treats memory: a trusted runtime below the agent charges each effect the loss expected to survive recovery, adds it to one running balance per principal, and then denies or escalates the next effect once that balance would exceed the authorized budget. This meets the three difficulties in turn: pricing residual loss rather than face value handles heterogeneity, charging below the agent stops a compromised model from misdeclaring or splitting an effect to slip under a cap, and a single shared balance turns a correlated burst into a visible overdraw that no per-effect gate can see. The outcome is that existing per-effect controls stop being independent gates and become spending policies over one shared resource. Realizing it takes a resource model that defines the quantity and the budgets it is charged against, plus a runtime that prices and reserves against them, both shown on the procurement example in Figure~\ref{fig:hero}.

We evaluate the abstraction with a controlled simulation of this procurement fleet, a study of public agent traces, and a ledger microbenchmark. The results support the claim and mark its limit. Local gates approve every purchase, yet the fleet overdraws its risk limit by $2.4\times$ on average, and by $48\times$ once it has grown to a thousand agents. The budget holds every run within the limit at every size, and because it prices effects by type it admits more useful work than a flat cap at the same safety. Its weakness is pricing: when effect types are misdeclared, or a shared trigger correlates losses that the charges assumed independent, realized loss can still exceed what the ledger accounted for. The ledger itself is cheap. Making those charges conservative and dependency-aware is the open problem.

The paper makes three contributions: irreversible exposure as a typed, cumulative admission quantity for agent operating systems (\S\ref{sec:resource}), a trusted runtime that combines risk-typed effects, hierarchical reserve-confirm-cancel ledgers, and value-at-risk admission control (\S\ref{sec:runtime}), and a controlled feasibility study that isolates the fleet-level composition failure and shows dependency-aware pricing to be the central open requirement (\S\ref{sec:study}).

\section{Background and Motivation}
\label{sec:background}
\paragraph{A running example} We make the procurement fleet concrete and reuse it throughout. Each of the fifty agents owns one product line and may place supplier orders up to \$50k under a per-agent rate limit, and the tenant tolerates \$250k of unhedged exposure over a trading day. On an ordinary day the agents buy independently and stay well inside that envelope. However, imagine a shared trigger where a supplier price-spike alert makes ``buy now'' rational for every agent at once. These are the parameters simulated by the evaluation in \S\ref{sec:study}.

\paragraph{Why it generalizes} This failure is not special to procurement. It appears wherever agents change the outside world: payroll runs, infrastructure automation, data deletion, customer communication, and incident response. Operators in these settings already cap scarce or risky quantities with quotas, approval limits, and desk-level risk controls, precisely because local permission has never implied aggregate safety. Agent fleets revive that pressure at higher speed and with weaker principal boundaries, which turns the operator's question from whether an action is allowed into how much irreversible exposure remains before the next commit.

\paragraph{Why counting is not enough} Answering that question means pricing effects rather than counting them. In the example, a canceled draft order, a completed supplier order, and a signed long-term contract carry very different residual loss once compensation, recovery, and revocation are exhausted. A control that treats every effect as one unit is therefore unsafe if it undercounts the irreversible ones and unusably conservative if it treats them all as maximal. Reversibility is a coordinate in the price rather than a binary gate.

\paragraph{The gap} Existing systems act one effect at a time. Table~\ref{tab:policies} lists the closest: Atomix defers settlement until it is safe~\cite{Mohammadi2026Atomix}, Cordon validates a task's authority before release~\cite{Chen2026Cordon}, actuarial gates such as AAI quote and bind capital for a single trajectory~\cite{Chen2026AAI}, guardrails and DLP validate payloads~\cite{OWASP2025AgenticThreats}, and sagas supply compensation paths~\cite{GarciaMolina1987Sagas,Helland2007LifeBeyond}. Each answers its local question well, and none keeps the cumulative, cross-agent balance that would have caught the fleet in the example. The missing mechanism is a pre-commit account that spans agents, workflows, and tenants.

\begin{table}[t]
\centering
\footnotesize
\caption{Existing mechanisms control local effects or domain-specific counters, but none maintains a heterogeneous cross-agent exposure account.}
\label{tab:policies}
\setlength{\tabcolsep}{4pt}
\begin{tabular}{@{}llll@{}}
\toprule
Mechanism & Question & Control & Scope \\
\midrule
Atomix~\cite{Mohammadi2026Atomix} & \emph{when} & defer settlement & per effect \\
Cordon~\cite{Chen2026Cordon} & \emph{whether} & validate authority & per task \\
AAI~\cite{Chen2026AAI} & \emph{how much} & quote and bind capital & per trajectory \\
Guardrails / DLP~\cite{OWASP2025AgenticThreats} & \emph{should pass} & validate payload & per call \\
Sagas / TCC~\cite{GarciaMolina1987Sagas,Helland2007LifeBeyond} & \emph{how to undo} & compensate or refund & per effect \\
Quotas / rate limits & \emph{how many} & cap a known counter & one unit \\
Pre-trade risk & \emph{how exposed} & cap position risk & finance \\
\bottomrule
\end{tabular}
\end{table}

\paragraph{Three requirements} The pre-commit account faces three difficulties, and each rules out a simpler design. Heterogeneous effects rule out a face-value collective cap, which charges a canceled draft and a signed contract alike and is safe only when it is uselessly conservative. Adversarial agents rule out trusting declared charges, because a compromised tool can split one large order into many sub-cap orders or misdeclare the effect type that sets the charge. Shared drivers rule out summing independent charges, since one price-spike alert or prompt injection makes many agents choose the same irreversible effect. A deployable control must therefore account before commit, canonicalize types, and price dependence.

\section{Resource Model}
\label{sec:resource}
\S\ref{sec:background} argued that a fleet needs a pre-commit account that prices effects, spans principals, and treats correlated exposures as more than a sum. This section defines that account as a resource. There are three parts: the charge on an effect, the budget a principal spends against, and the risk-pressure signal that drives admission.

\paragraph{Irreversible exposure} For an external effect $e$, define its charge $c(e)$ as the residual loss over a configured horizon after available compensation, recovery, and revocation have been applied. Effect outcomes are stochastic, so the runtime charges at a confidence level: $c_\gamma(e)$ is the $\gamma$-quantile of the residual-loss distribution supplied by the pricing component. The charge is a scalar in a financial domain, or a vector when money, data loss, availability, and legal exposure cannot be reduced to a common unit. Reversibility~\cite{GarciaMolina1987Sagas} enters the price as one coordinate among several rather than acting as a binary gate.

\paragraph{Irreversibility budget} Each principal $p$ is assigned a windowed budget $B_p$, and the runtime maintains a balance of reserved plus committed exposure against it. The invariant is simple: a principal may not externalize an effect when the marginal charge would exceed its budget at the required confidence level. In the running example, each of the fifty agents spends against its own allocation, its workflow's, and the tenant's \$250k envelope, so no agent's local freedom can overdraw the fleet's ceiling. Budgets are replenished on a horizon set by the budget authority: short windows bound bursts, longer windows bound campaigns.

\paragraph{Risk pressure} We call the depletion signal \emph{risk pressure}, the analog of memory pressure: as a kernel reclaims pages and throttles when free memory runs low, the runtime tightens admission as the free budget falls and, in the extreme, freezes externalization (\S\ref{sec:runtime}). Risk pressure is a scheduling signal rather than a proof of safety, turning a fixed per-call verdict into a control loop over a scarce quantity.

\section{The Runtime}
\label{sec:runtime}
The resource model says what to charge, and the runtime is the machinery that charges it. Its jobs are to make an effect's exposure observable before the effect commits, to stop a fleet from double-spending one budget, and to decide admission when the budget runs low. The mechanisms below meet these jobs, each an existing primitive moved below the agent's trust boundary, so a compromised model can neither price nor quietly spend its own risk.

\paragraph{Risk-typed effects} To make exposure observable before commit, tool specifications declare effect classes, exposure bounds, reversibility paths, compensation mechanisms, authority requirements, and dependency hints. Existing taxonomies of reversible, compensatable, and irreversible effects provide the initial type structure~\cite{GarciaMolina1987Sagas}. Pricing sits below the agent, so charges $c_\gamma(e)$ come from specifications and a trusted pricing service, never from the model; the charge itself is only as trusted as the specification author and its auditor. Many domains already encode partial versions of this information in approval rules, cloud quotas, and retention policies, and the runtime makes those implicit controls executable at the commit boundary.

\paragraph{Canonicalization} A charge is only as sound as the type it is read from, so the runtime assigns that type itself rather than accepting one from the caller. A trusted registry binds each effect class to the evidence that identifies it: the calling tool's identity and signed specification, the API endpoint and request schema the call resolves to, and the settlement receipt that later confirms what actually happened. A compromised workflow can therefore propose a final transfer, but it cannot declare that transfer refundable, because it never supplies the label. What the registry cannot resolve is charged at the most expensive matching class, so an unrecognized effect is throttled rather than silently admitted. The trusted base is wider than the ledger: the registry and its auditors, the tool wrappers, the pricing service, and principal attribution.

\paragraph{Charge lifecycle} Charges are initialized from the conservative bound in the effect specification, calibrated offline by replaying traces and stress cases against realized residual loss, and audited against settlement receipts, which are the only ground truth about how much recovery actually returned. They are adjusted when the pricer detects a dependency, either a correlation class declared on a workflow or a shared trigger inferred from the proposal stream, so the marginal charge for the next effect in that class rises instead of repeating the independent price. \S\ref{sec:study} breaks both the binding and the calibration, and reports what each is worth.

\paragraph{Hierarchical ledgers and reservations} Per-principal ledgers expose native \texttt{reserve}, \texttt{confirm}, and \texttt{cancel} calls that keep a fleet from double-spending one risk envelope: speculative branches reserve before acting, confirm on commit, and refund on cancellation or successful compensation~\cite{ONeil1986Escrow,Helland2007LifeBeyond}. Admission checks every ledger on the effect's principal path and reserves atomically across them. Algorithm~\ref{alg:admit} gives the reserve procedure. The ledger must guarantee four invariants: no overdraft, idempotent confirmation, crash recovery, and hierarchical roll-up.

\begin{algorithm}[t]
\footnotesize
\caption{Value-at-risk admission (reserve).}
\label{alg:admit}
\begin{algorithmic}[1]
\State \textbf{input:} effect $e$, principal path $P$ (agent $\to$ workflow $\to$ tenant)
\State $c \gets \textsc{pricer.charge}(e,\ \text{context},\ \gamma)$ \Comment{charge, never from the agent}
\State \textbf{atomically}
\ForAll{ledger $\ell \in P$}
  \If{$\ell.\text{reserved} + \ell.\text{committed} + c > \ell.B$} \Return \textsc{deny} \EndIf
\EndFor
\ForAll{ledger $\ell \in P$} $\ell.\text{reserved} \mathrel{+}= c$ \EndFor
\State \Return reservation id \Comment{\textsc{confirm} commits it; \textsc{cancel} refunds it}
\end{algorithmic}
\end{algorithm}

\paragraph{Admission and scheduling under risk pressure} \emph{Value-at-risk admission control} is the rule: an effect commits only if reserved and committed exposure stay within allocation at every principal on the agent-to-workflow-to-tenant path. As risk pressure rises, the runtime tightens confidence, demotes low-value spenders, expires stale reservations, batches approvals, or escalates to a rate-limited authority with explicit latency and capacity budgets. Escalation does not raise $B$. Instead, the authority mints a separate audited exception allocation and charges the effect against that, so the principal's ordinary budget is untouched and the extra exposure is recorded as approved rather than as headroom, because approved exposure is risk a principal chose to take, whereas the failure mode is risk the ledger failed to stop. Delegation treats \emph{capabilities as risk currency}: attenuable tokens denominated in risk units that an agent must spend rather than merely present. Schedulers already trade fairness, priority, and utilization under scarcity; here the scarce quantity is irreversible exposure, and denial happens before the effect commits.

\paragraph{The contract} The ledger guarantee is narrow. It prevents overdraft in \emph{declared charge units}, under four assumptions: authenticated principals, strongly consistent budget authorities, idempotent reservation lifecycles, and charges supplied by trusted effect specifications. It does not guarantee bounded realized loss under stale prices, correlations absent from the charge model, malicious type declarations, or misattributed workflows. The design therefore exposes two layers of guarantee: a \emph{declared-charge bound} that the ledger enforces unconditionally, and a \emph{realized-loss bound} that depends on calibrated, dependency-aware pricing.

\section{Evaluation}
\label{sec:study}
We evaluate the runtime design as a controlled feasibility study rather than a deployed agent OS. Five research questions organize it: whether per-effect safety fails under composition and a budgeted ledger can bound exposure before commit (RQ1), whether the need for accounting persists under fleet growth, fragmentation, and reactive alternatives (RQ2), what liveness and scheduling costs the budget introduces (RQ3), when risk typing and calibration help and when they fail (RQ4), and whether public agent traces support the shared-trigger correlation the simulation assumes (RQ5).

\paragraph{Workload} The main workload is a discrete-event procurement simulation of the running example, and Table~\ref{tab:params} lists its parameters. Routine demand typically fills roughly 85\% of the budget, a correlated market trigger raises every agent's proposal probability for 50 ticks, and an executed purchase of value $v$ realizes residual exposure $v(1-r)$. Charges come from declared effect types, and only the ledger microbenchmark models admission latency.

\paragraph{Baselines} The \emph{local-gates} baseline enforces a per-call cap and per-agent rate limit with no shared state, and the \emph{budget} runtime reserves the 95th-percentile charge $c_{0.95}(e)=0.52\,v$ and denies any effect whose charge would push the reserved balance past $B$. We also run the strongest aggregate alternatives an operator would use: face-value and pooled-quantile caps, static per-agent partitions, a reactive circuit breaker, and hierarchical sub-budgets.

\paragraph{Metrics and protocol} A run \emph{overdraws} when realized exposure that no authority approved exceeds $R$ in any window, so overdraw measures what the ledger let through silently. We also report realized exposure relative to $R$, admitted routine and burst traffic, attacker value moved, sibling-workflow throughput, and reservation latency. Results are means over 300 seeded runs with 95\% confidence intervals and Wilson intervals for overdraw proportions. We release the simulator, seeds, parameters, benchmark, and trace analysis at \url{https://github.com/mpi-dsg/irreversibility-budget}.

\paragraph{Ledger overhead} The ledger microbenchmark measures the reservation lifecycle in isolation rather than an integrated runtime, on a single host, in memory, with no persistence, replication, or crash recovery. A reserve-then-confirm across a three-level agent-workflow-tenant path costs $2.6\,\mu$s at the median and 240 bytes per live reservation, and a shared tenant root under 32 concurrent threads sustains a few $\times10^5$ reservation cycles per second. This number bounds the accounting work rather than the cost of a durable budget authority, which must add a log write on every spend and, across hosts, a round of coordination on any violating spend.

\begin{table}[t]
\footnotesize
\caption{Simulation parameters.}
\label{tab:params}
\setlength{\tabcolsep}{4pt}
\begin{tabular}{@{}ll@{}}
\toprule
Parameter & Value \\
\midrule
Fleet / window & 50 agents / 1{,}000 ticks \\
Proposal prob.\ (calm, burst) & $0.0005$; $0.05$ for 50 ticks \\
Purchase value & log-normal, median $\sim$\$15k \\
Recovery fraction $r$ & $\mathrm{Beta}(6,2)$, mean $0.75$ \\
Tolerance $R$ / budget $B$ & \$250k / $R$ \\
Charge $c_{0.95}(e)$ & $0.52\,v$ (95th-pct.\ residual) \\
Local baseline & \$50k call cap $+$ rate limit \\
\bottomrule
\end{tabular}
\end{table}

\subsection{Composition (RQ1)}
We ask whether per-effect safety composes to the fleet. Every proposal stays within its per-call cap and rate limit, so the only difference between arms is whether admission also checks an aggregate budget.

\begin{figure}[t]
\centering
\includegraphics[width=0.84\columnwidth]{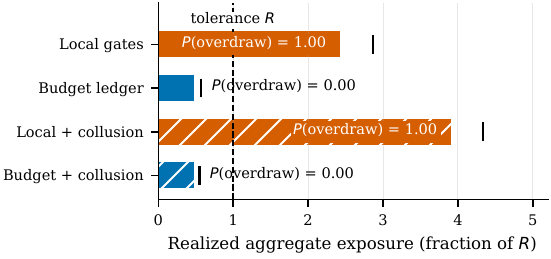}
\caption{Realized aggregate exposure as a fraction of tolerance $R$, with and without a colluding fragmentation attack. Bars are means over 300 runs; ticks mark the 95th percentile. Local gates approve every action and overdraw in every run. The budget bounds exposure below $R$ in both settings.}
\Description{Horizontal bar chart of realized aggregate exposure relative to tolerance for four configurations; local gates exceed tolerance, budget stays below it.}
\label{fig:e1}
\end{figure}

\paragraph{Results} Figure~\ref{fig:e1} contrasts the two arms. Local gates approve every purchase and still overdraw in all 300 runs, at $2.4\times$ tolerance on average, while the budget overdraws in none of them, with a 95\% upper bound of 1.3\% on overdraw probability, and settles near $0.48\times$ tolerance. The gap is structural rather than a tuning artifact: no per-call gate can see the running sum, so none can deny the marginal purchase that tips the fleet over. The budget's $0.48\times$ occupancy is simply the arithmetic of a full ledger whose realized exposure concentrates near the mean-to-quantile ratio $0.25/0.52$ of $B$.

\subsection{Scale and alternatives (RQ2)}
We ask whether the gap grows, and whether cheaper alternatives close it. We hold $R$ fixed and grow the fleet from ten to a thousand agents, then add two variants: a fragmentation attack that splits one economic action into many sub-cap ones, and a circuit breaker that halts the fleet once observed loss reaches $R$.

\paragraph{Results} The overdraw grows with the fleet. Local gates are safe at ten agents but reach $2.4\times$ tolerance at fifty, $9.7\times$ at two hundred, and $48\times$ at a thousand, because aggregate demand scales while each agent individually stays compliant. The budget holds near $0.48\times$ at every size. The alternatives fail for instructive reasons. Fragmentation slips the full \$1.5M transfer past local gates to $3.9\times$ tolerance, yet the ledger admits only \$459k of that value, however finely the attacker splits it, because the charge attaches to exposure rather than to call count. The circuit breaker ends at $1.03\times R$ with instant observation and $2.2\times R$ at a fifty-tick lag, because it reads loss only after the irreversible effect has already committed.

\paragraph{Takeaway} The gap widens with scale, and only a control that reserves before commit, rather than counting calls or reacting to loss, stays safe as the fleet grows.

\subsection{Liveness (RQ3)}
Safety by denial is easy to state and easy to abuse, so we ask what the budget costs in useful work. We sweep the budget size $B/R$. Separately, four scheduling arms stress hoarding, sibling isolation, campaign pacing, and re-authorization.

\begin{figure}[t]
\centering
\includegraphics[width=0.84\columnwidth]{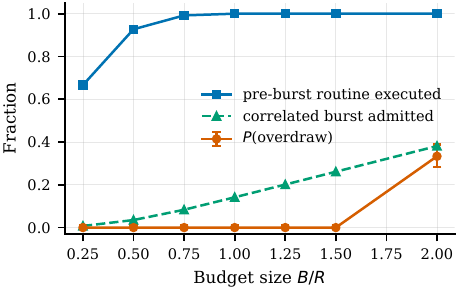}
\caption{Safety and liveness against budget size $B/R$ (300 runs; error bars are 95\% Wilson intervals on the overdraw proportion). Larger budgets admit more burst traffic and, past $1.5R$, increase overdraw risk.}
\Description{Line chart of overdraw probability, routine traffic executed, and burst admitted as budget size grows.}
\label{fig:e2}
\end{figure}

\paragraph{Results} Figure~\ref{fig:e2} shows the trade-off. A budget of $B=R$ passes routine pre-burst traffic in full and admits 14\% of the correlated burst, and even $B=R/2$ preserves 93\% of routine traffic, so the price of safety falls mostly on the burst rather than on ordinary work. The sharper cost is temporal: once the window's budget is spent, post-burst traffic starves until replenishment. Enlarging the budget buys burst headroom but reintroduces overdraw, up to 33\% of runs at $B=2R$, so the knob is real but not free. Table~\ref{tab:secondary} shows the scheduling arms recovering most of the lost liveness: fair-share admission curbs hoarding, per-workflow ledgers lift an innocent sibling's throughput from 48\% to 89\%, and rate-limited re-authorization restores post-burst liveness to 93\% while making the extra \$75k of exposure audited rather than silent.

\paragraph{Takeaway} The budget trades burst liveness for safety on a tunable knob, and per-workflow ledgers plus paced re-authorization recover most of it without hiding exposure.

\begin{table}[t]
\footnotesize
\caption{Four secondary arms: the budget's scheduling policies and liveness stress tests, 300 runs each.}
\label{tab:secondary}
\setlength{\tabcolsep}{3pt}
\begin{tabular}{@{}p{0.20\columnwidth}p{0.30\columnwidth}p{0.42\columnwidth}@{}}
\toprule
Arm & Setup & Result \\
\midrule
Fair-share & per-principal cap $0.05B$, fragmentation & attacker value moved \$459k $\to$ \$200k; bound holds against 40 colluders \\
Replenishment & three windows, paced campaign & per-window resets admit 72\% of a campaign; a flat $1.5R$ horizon ledger is an outage or a no-op \\
Hierarchy & rogue workflow, flat vs.\ nested ledger & sibling throughput 48\% $\to$ 89\% with a $0.6B$ sub-budget; tenant bound holds \\
Re-authorization & rate-limited authority, $H$ effects/window & $H{=}50$ restores post-burst liveness to 93\%, adding \$75k of audited exposure \\
\bottomrule
\end{tabular}
\end{table}

\subsection{Typing and pricing (RQ4)}
The ledger enforces only declared charges, so we ask whether typing effects by residual loss beats flat counting, and how charge errors actually fail. We scale declared charges by a factor $\varepsilon$ around the true quantile, compare a typed ledger against face-value and pooled caps on a mixed refundable-and-final workload, and then break the charges' independence assumption with a burst that depresses recovery.

\begin{figure}[t]
\centering
\includegraphics[width=0.84\columnwidth]{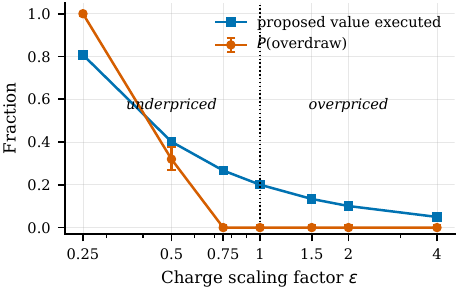}
\caption{Mispricing sensitivity: declared charges scaled by $\varepsilon$ relative to the 95th-percentile charge (300 runs; error bars are 95\% Wilson intervals). Underpricing fails toward silent overdraw; overpricing toward throttling.}
\Description{Line chart of overdraw probability and executed value as declared charges are scaled below and above the true quantile.}
\label{fig:e3}
\end{figure}

\paragraph{Results} Figure~\ref{fig:e3} shows that mispricing is asymmetric. Underpricing fails silently, overdrawing in 32\% of runs already at half the true charge, whereas overpricing fails only as throttling ($\varepsilon=4$ executes 5\% of proposed value), a failure the scheduler can see and correct; charges should therefore fail conservatively. When charges are right, typing pays: the typed ledger executes $1.8\times$ the value of a face-value cap and $1.7\times$ the strongest pooled cap, a gain that tracks heterogeneity and shrinks to $1.2\times$ as effect classes converge. Two failures bound that result. Misdeclaring colluding transfers as refundable re-opens the cap to \$1.0M and overdraws in 77\% of runs, and a shared trigger that depresses recovery from $0.75$ to $0.25$ makes additive per-effect charging overdraw in 59\% of runs while the ledger believes itself safe. Repricing at the burst quantile fixes the latter only with near-instant detection: a five-tick lag in a fifty-tick burst still overdraws 39\% of runs, and a ten-tick lag is no better than no detector at all.

\paragraph{Takeaway} Typing buys real utility when charges are honest, but misdeclaration and correlated recovery let realized loss outrun the ledger's belief, which makes conservative, dependency-aware pricing the central open requirement.

\subsection{Trace evidence (RQ5)}
The whole model rests on shared triggers correlating losses, which the simulator cannot justify on its own, so we test it against real behavior. We analyzed 38{,}452 trajectories from the public $\tau$-bench~\cite{Yao2024TauBench} and AgentDojo~\cite{Debenedetti2024AgentDojo} benchmarks and asked whether irreversible effects are agent-idiosyncratic or task-determined. In $\tau$-bench we read the 165 tasks with at least three trials as fleets of independent agents meeting the same context, and compare their agreement against a permutation null that breaks the task-to-effect link while preserving each effect's marginal rate. In AgentDojo the runs sharing one planted prompt injection, a single instruction placed in a shared resource and met by agents running different user tasks, are the fleet; we report the 35 injections with at least 20 runs. A fixed vocabulary map tags pay, refund, delete, update, send, and create as external effects, and marks a prefix read-only only when every call in it unambiguously is.

\paragraph{Results} External effects account for 22\% of tool calls, and a single agent's effects are typically sparse and end-loaded, so no agent bursts on its own. The correlation lives across agents: which external effect fires is task-determined rather than agent-specific ($z$ up to $198$ against an independent-agent null), and one planted instruction propagates the same external effect across 84\% of a heterogeneous fleet on average (minimum 48\%). This is precisely the structure the additive charges in RQ4 fail to price.

\paragraph{Takeaway} Real traces confirm that shared context drives correlated external effects, so the regime the budget targets is the common case.

\section{Discussion and Limitations}
\label{sec:open}
\paragraph{Limitations} The study establishes the accounting mechanism rather than a deployed system. It does not show that value-at-risk is the right risk measure, that calibration is solved, or that distributed ledgers stay available under partition, and it does not exercise the full reserve-confirm-cancel lifecycle, capability delegation, vector charges, or refunds. Two biases follow: liveness cost is measured pessimistically because refunds are absent, and safety is measured optimistically because prices are constructed rather than learned. These failures define the contract a deployable budget must meet: trusted effect typing, dependency-aware charging, workflow-level attribution, and denial before commit.

\emph{Pricing and budget setting.} The budget is only as sound as its charges and as useful as the authority's choice of $B$. A business tolerance does not directly define a charge budget, and larger budgets trade liveness for weaker realized-loss bounds. Vector budgets may be needed when losses are not commensurable, since data exposure, availability damage, and reputational harm resist a common scalar in a way money does not. We use value-at-risk because operators know it and it is cheap to simulate, but the abstraction does not require it, and value-at-risk is not subadditive while expected shortfall is coherent~\cite{Artzner1999Coherent}. Charges should fail conservatively, so mispricing escalates rather than silently admits.

\emph{Correlation.} The tractable middle ground between independent charges and worst-case coupling remains open: correlation classes declared per workflow, shared-trigger detection from proposal streams, and marginal charges under an explicit dependency model. Detection must run early enough to affect admission before most of the burst's exposure commits.

\emph{Distributed accounting.} A budget that can be double-spent provides telemetry rather than control. Across hosts, hard budgets require coordination on any violating spend, while advisory budgets reconcile later and cannot enforce safety. The runtime should separate the two, shard by principal hierarchy, and report the availability cost of strong budget authorities.

\emph{Adversarial spenders.} Prompt-injected or colluding agents can fragment actions below thresholds, hoard reservations, misdeclare types, or manipulate refunds~\cite{Debenedetti2024AgentDojo,OWASP2025AgenticThreats}. Canonicalization must therefore span agents and workflows, because the shared budget itself becomes the target, and re-authorization must be rate-limited and audited rather than an unlimited human escape hatch. Our bonded-recourse design settles the compensations themselves~\cite{Bindschaedler2026Recourse}.

\emph{Liveness.} A budget converts some unsafe commits into denial, contention, and auditable re-authorization. It can hold the bound and still starve routine work after a burst, or let an attacker convert theft into reservation denial-of-service. Reservation expiry, workflow sub-budgets, fair-share admission, replenishment horizons, and authority overrides are scheduling policy.

\section{Related Work}
\label{sec:related}
This work combines existing mechanisms from database transactions, OS resource accounting, financial risk control, and agent safety. The novelty is their use together as a cross-principal admission resource for typed, heterogeneous, adversarial agent effects.

\paragraph{Database and OS primitives.} Database escrow~\cite{ONeil1986Escrow} and saga compensation~\cite{GarciaMolina1987Sagas,Helland2007LifeBeyond} supply the reserve-confirm-cancel lifecycle, resource containers~\cite{Banga1999ResourceContainers} supply the move of charging a principal rather than a process, and coordination avoidance~\cite{Hellerstein2020CALM,Bailis2014Coordination} separates invariant-preserving local updates from updates that require agreement. We apply these primitives to risk units at the tool-commit boundary, where the runtime must be able to deny before an external effect settles.

\paragraph{Agent effect controls.} The mechanisms summarized in Table~\ref{tab:policies} protect individual effects or branches by controlling settlement time, semantic authority, or payload validity~\cite{Mohammadi2026Atomix,Chen2026Cordon,OWASP2025AgenticThreats}. Our group's vision papers argue for OS-level confinement of exploratory effects~\cite{Hu2026ExternalizationBarriers} and for checking effects instead of trusting code~\cite{Hu2026DontTrust}; the budget adds what these lack: cumulative accounting across agents, workflows, and tenants.

\paragraph{Risk pricing and safe learning.} Settlement and underwriting systems price or compensate single actions and trajectories~\cite{Hua2026ARS,Xu2026TraceEconomic}. Constrained and safe reinforcement learning caps cumulative safety cost inside one agent's policy optimization~\cite{Achiam2017CPO}. Single-agent actuarial admission gates price actions and defend against fragmentation within a trajectory~\cite{Chen2026AAI}. Our setting moves pricing below untrusted agents and aggregates across principals, so collusion and shared drivers naturally become first-order concerns.

\paragraph{Governance and operational caps.} A governance proposal already names an \emph{irreversibility budget}, but treats it as an additive cap with human re-authorization~\cite{Sahoo2026Controllability}. Additivity suits sparse, human-supervised settings rather than dense fleets with fragmentation, misdeclaration, and correlated recoverability. Credit-card limits, cloud quotas, and pre-trade checks cap one known unit, whereas agent fleets need typed residual-loss accounting. The budget also differs from access control, which decides whether an action is allowed: the budget denies an allowed action because the aggregate would overdraw, a claim no per-call mechanism expresses.

\section{Conclusion}
No runtime tracks the irreversible exposure that fleets of tool-using agents build up together. We proposed the irreversibility budget, a cumulative per-principal account that a trusted runtime charges before commit, denying an effect once the fleet would overdraw. Local gates approved every action and still overdrew by up to $48\times$, while the budget held every correctly charged run within the risk limit. Conservative, dependency-aware pricing remains open, but irreversibility as a first-class resource gives agent operating systems a place to account for it.

\balance
\bibliographystyle{ACM-Reference-Format}
\bibliography{main}

@inproceedings{Mohammadi2026Atomix,
  author    = {Mohammadi, Bardia and Potamitis, Nearchos and Klein, Lars Henning and Arora, Akhil and Bindschaedler, Laurent},
  title     = {{Atomix}: Timely, Transactional Tool Use for Reliable Agentic Workflows},
  booktitle = {{ICLR} 2026 Workshop on Agents in the Wild: Safety, Security, and Beyond ({AIWILD})},
  year      = {2026},
  publisher = {OpenReview.net},
  address   = {Rio de Janeiro, Brazil},
  url       = {https://openreview.net/forum?id=UeRbEpSVUz},
  note      = {Also available as arXiv:2602.14849 [cs.LG]}
}

@misc{Chen2026Cordon,
  author        = {Chen, Zheng and Liu, Hanqing and Xu, Duling and Dong, Dong and Li, Jialin and Pu, Bangzheng and Zhai, Jidong},
  title         = {{Cordon}: Semantic Transactions for Tool-Using {LLM} Agents},
  year          = {2026},
  eprint        = {2606.17573},
  archivePrefix = {arXiv},
  primaryClass  = {cs.OS},
  doi           = {10.48550/arXiv.2606.17573},
  url           = {https://arxiv.org/abs/2606.17573}
}

@inproceedings{GarciaMolina1987Sagas,
  author    = {Garcia-Molina, Hector and Salem, Kenneth},
  title     = {Sagas},
  booktitle = {Proceedings of the 1987 {ACM} {SIGMOD} International Conference on Management of Data ({SIGMOD} '87)},
  year      = {1987},
  pages     = {249--259},
  publisher = {ACM Press},
  address   = {New York, NY, USA},
  location  = {San Francisco, CA, USA},
  doi       = {10.1145/38713.38742}
}

@inproceedings{Helland2007LifeBeyond,
  author    = {Helland, Pat},
  title     = {Life beyond Distributed Transactions: an Apostate's Opinion},
  booktitle = {Proceedings of the Third Biennial Conference on Innovative Data Systems Research ({CIDR} 2007)},
  year      = {2007},
  month     = jan,
  pages     = {132--141},
  publisher = {CIDR},
  address   = {Asilomar, CA, USA},
  url       = {https://www.cidrdb.org/cidr2007/papers/cidr07p15.pdf}
}

@article{ONeil1986Escrow,
  author    = {O'Neil, Patrick E.},
  title     = {The {Escrow} Transactional Method},
  journal   = {{ACM} Transactions on Database Systems},
  volume    = {11},
  number    = {4},
  pages     = {405--430},
  year      = {1986},
  month     = dec,
  publisher = {Association for Computing Machinery},
  doi       = {10.1145/7239.7265}
}

@article{Hellerstein2020CALM,
  author    = {Hellerstein, Joseph M. and Alvaro, Peter},
  title     = {Keeping {CALM}: When Distributed Consistency Is Easy},
  journal   = {Communications of the {ACM}},
  volume    = {63},
  number    = {9},
  pages     = {72--81},
  year      = {2020},
  month     = sep,
  publisher = {Association for Computing Machinery},
  doi       = {10.1145/3369736}
}

@article{Bailis2014Coordination,
  author    = {Bailis, Peter and Fekete, Alan D. and Franklin, Michael J. and Ghodsi, Ali and Hellerstein, Joseph M. and Stoica, Ion},
  title     = {Coordination Avoidance in Database Systems},
  journal   = {Proceedings of the {VLDB} Endowment},
  volume    = {8},
  number    = {3},
  pages     = {185--196},
  year      = {2014},
  month     = nov,
  publisher = {{VLDB} Endowment},
  doi       = {10.14778/2735508.2735509}
}

@misc{Xu2026TraceEconomic,
  author        = {Xu, Binyan and Dai, Xilin and Yang, Fan and Zhang, Kehuan},
  title         = {When Agent Automation Becomes Profitable: Quantifying and Insuring Autonomous {AI} Risk through Trace-Economic Underwriting},
  year          = {2026},
  eprint        = {2606.16465},
  archivePrefix = {arXiv},
  primaryClass  = {cs.AI},
  doi           = {10.48550/arXiv.2606.16465},
  url           = {https://arxiv.org/abs/2606.16465}
}

@inproceedings{Sahoo2026Controllability,
  author    = {Sahoo, Subramanyam},
  title     = {The Controllability Trap: A Governance Framework for Military {AI} Agents},
  booktitle = {{ICLR} 2026 Workshop on Agents in the Wild: Safety, Security, and Beyond ({AIWILD})},
  year      = {2026},
  publisher = {OpenReview.net},
  address   = {Rio de Janeiro, Brazil},
  url       = {https://openreview.net/forum?id=yLkGf4Uw14},
  note      = {Also available as arXiv:2603.03515 [cs.CY]}
}

@inproceedings{Madras2018LearningToDefer,
  author    = {Madras, David and Pitassi, Toniann and Zemel, Richard S.},
  title     = {Predict Responsibly: Improving Fairness and Accuracy by Learning to Defer},
  booktitle = {Advances in Neural Information Processing Systems 31 ({NeurIPS} 2018)},
  volume    = {31},
  pages     = {6150--6160},
  year      = {2018},
  publisher = {Curran Associates, Inc.},
  address   = {Montreal, Canada}
}

@inproceedings{Mozannar2020LearningToDefer,
  author    = {Mozannar, Hussein and Sontag, David},
  title     = {Consistent Estimators for Learning to Defer to an Expert},
  booktitle = {Proceedings of the 37th International Conference on Machine Learning ({ICML} 2020)},
  series    = {Proceedings of Machine Learning Research},
  volume    = {119},
  pages     = {7076--7087},
  year      = {2020},
  month     = {13--18 Jul},
  editor    = {Daum{\'e} III, Hal and Singh, Aarti},
  publisher = {PMLR},
  address   = {Virtual Event},
  url       = {https://proceedings.mlr.press/v119/mozannar20b.html}
}

@inproceedings{Debenedetti2024AgentDojo,
  author    = {Debenedetti, Edoardo and Zhang, Jie and Balunovi{\'c}, Mislav and Beurer-Kellner, Luca and Fischer, Marc and Tram{\`e}r, Florian},
  title     = {{AgentDojo}: A Dynamic Environment to Evaluate Prompt Injection Attacks and Defenses for {LLM} Agents},
  booktitle = {Advances in Neural Information Processing Systems 37 ({NeurIPS} 2024)},
  year      = {2024},
  volume    = {37},
  pages     = {82895--82920},
  publisher = {Curran Associates, Inc.},
  address   = {Vancouver, Canada},
  doi       = {10.52202/079017-2636},
  note      = {Datasets and Benchmarks Track}
}

@misc{Packer2023MemGPT,
  author        = {Packer, Charles and Wooders, Sarah and Lin, Kevin and Fang, Vivian and Patil, Shishir G. and Stoica, Ion and Gonzalez, Joseph E.},
  title         = {{MemGPT}: Towards {LLMs} as Operating Systems},
  year          = {2023},
  eprint        = {2310.08560},
  archivePrefix = {arXiv},
  primaryClass  = {cs.AI},
  doi           = {10.48550/arXiv.2310.08560},
  url           = {https://arxiv.org/abs/2310.08560}
}

@inproceedings{Mei2025AIOS,
  author    = {Mei, Kai and Zhu, Xi and Xu, Wujiang and Jin, Mingyu and Hua, Wenyue and Li, Zelong and Xu, Shuyuan and Ye, Ruosong and Ge, Yingqiang and Zhang, Yongfeng},
  title     = {{AIOS}: {LLM} Agent Operating System},
  booktitle = {Second Conference on Language Modeling ({COLM} 2025)},
  year      = {2025},
  publisher = {OpenReview.net},
  address   = {Montreal, Canada},
  url       = {https://openreview.net/forum?id=L4HHkCDz2x},
  note      = {Also available as arXiv:2403.16971}
}

@misc{OWASP2025AgenticThreats,
  author       = {{OWASP Gen AI Security Project, Agentic Security Initiative}},
  title        = {{Agentic AI} - Threats and Mitigations},
  year         = {2025},
  month        = feb,
  howpublished = {White paper, version 1.1},
  url          = {https://genai.owasp.org/resource/agentic-ai-threats-and-mitigations/},
  note         = {Version 1.1, February 2025}
}

@article{Artzner1999Coherent,
  author    = {Artzner, Philippe and Delbaen, Freddy and Eber, Jean-Marc and Heath, David},
  title     = {Coherent Measures of Risk},
  journal   = {Mathematical Finance},
  volume    = {9},
  number    = {3},
  pages     = {203--228},
  year      = {1999},
  month     = jul,
  publisher = {Wiley},
  doi       = {10.1111/1467-9965.00068}
}

@misc{Chen2026AAI,
  author        = {Chen, Hao-Hsuan},
  title         = {Insuring Every Action: An Authority Frontier Framework for Runtime Actuarial Control of Autonomous {AI} Agents},
  year          = {2026},
  eprint        = {2605.25632},
  archivePrefix = {arXiv},
  primaryClass  = {cs.AI},
  doi           = {10.48550/arXiv.2605.25632},
  url           = {https://arxiv.org/abs/2605.25632}
}

@misc{Hua2026ARS,
  author        = {Hua, Wenyue and Peng, Tianyi and Wang, Chi and Pei, Jiaxin and Kaufman, Ian and Lim, Bryan and Fang, Chandler},
  title         = {Quantifying Trust: Financial Risk Management for Trustworthy {AI} Agents},
  year          = {2026},
  eprint        = {2604.03976},
  archivePrefix = {arXiv},
  primaryClass  = {cs.AI},
  doi           = {10.48550/arXiv.2604.03976},
  url           = {https://arxiv.org/abs/2604.03976}
}

@inproceedings{Banga1999ResourceContainers,
  author    = {Banga, Gaurav and Druschel, Peter and Mogul, Jeffrey C.},
  title     = {Resource Containers: A New Facility for Resource Management in Server Systems},
  booktitle = {Proceedings of the Third {USENIX} Symposium on Operating Systems Design and Implementation ({OSDI} '99)},
  year      = {1999},
  month     = feb,
  pages     = {45--58},
  publisher = {USENIX Association},
  address   = {New Orleans, LA, USA},
  url       = {https://www.usenix.org/conference/osdi-99/resource-containers-new-facility-resource-management-server-systems}
}

@inproceedings{Achiam2017CPO,
  author    = {Achiam, Joshua and Held, David and Tamar, Aviv and Abbeel, Pieter},
  title     = {Constrained Policy Optimization},
  booktitle = {Proceedings of the 34th International Conference on Machine Learning ({ICML} 2017)},
  series    = {Proceedings of Machine Learning Research},
  volume    = {70},
  pages     = {22--31},
  year      = {2017},
  month     = {06--11 Aug},
  editor    = {Precup, Doina and Teh, Yee Whye},
  publisher = {PMLR},
  address   = {Sydney, Australia},
  url       = {https://proceedings.mlr.press/v70/achiam17a.html}
}

@inproceedings{Yao2024TauBench,
  author    = {Yao, Shunyu and Shinn, Noah and Razavi, Pedram and Narasimhan, Karthik},
  title     = {{$\tau$-bench}: A Benchmark for Tool-Agent-User Interaction in Real-World Domains},
  booktitle = {The Thirteenth International Conference on Learning Representations ({ICLR} 2025)},
  year      = {2025},
  publisher = {OpenReview.net},
  address   = {Virtual Event},
  url       = {https://openreview.net/forum?id=roNSXZpUDN},
  note      = {Also available as arXiv:2406.12045}
}

@inproceedings{Hu2026ExternalizationBarriers,
  author    = {Hu, Jinhao and Mohammadi, Bardia and Goel, Ashvin and Bindschaedler, Laurent},
  title     = {Externalization Barriers: An {OS} Abstraction for Untrusted Agent Exploration},
  booktitle = {The 2nd Workshop on {OS} Design for {AI} Agents ({AgenticOS} 2026)},
  year      = {2026},
  publisher = {ACM},
  address   = {Prague, Czech Republic},
  note      = {To appear}
}

@inproceedings{Hu2026DontTrust,
  author    = {Hu, Jinhao and Goel, Ashvin and Bindschaedler, Laurent},
  title     = {Don't Trust the Code, Check Its Effects},
  booktitle = {Proceedings of the 5th Workshop on Practical Adoption Challenges of {ML} for Systems ({PACMI} 2026)},
  year      = {2026},
  publisher = {ACM},
  note      = {To appear}
}

@inproceedings{Bindschaedler2026Recourse,
  author    = {Bindschaedler, Laurent and Botha, Quentin and Siebenbrunner, Christoph},
  title     = {Bonded Recourse for Smart-Contract Settlement of Compensable Agent Side Effects},
  booktitle = {Proceedings of the 8th International Conference on Blockchain Computing and Applications ({BCCA} 2026)},
  year      = {2026},
  publisher = {IEEE},
  address   = {Barcelona, Spain},
  note      = {To appear}
}

\end{document}